\pdfoutput=1
\documentclass[a4paper]{article}
\usepackage{interspeech2010,amssymb,amsmath,epsfig}
\usepackage[utf8]{inputenc}
\usepackage{listings}
\usepackage[activate]{microtype}
\ninept
\def\reg{{\rm\ooalign{\hfil
     \raise.07ex\hbox{\scriptsize R}\hfil\crcr\mathhexbox20D}}}

\title{A Broadcast News Corpus for Evaluation and Tuning of \\German LVCSR Systems}

\makeatletter
\def\name#1{\gdef\@name{#1\\}}
\makeatother
\name{{\em Felix Weninger$^1$, Bj\"orn Schuller$^1$, Florian Eyben$^1$, Martin W\"ollmer$^1$, and Gerhard Rigoll$^1$}}

\address{$^1$Institute for Human-Machine Communication, Technische Universit\"at M\"unchen, Germany \\
{\small \tt \{weninger|schuller|eyben|woellmer|rigoll\}@tum.de}}

\begin{document}
\maketitle
\begin{abstract}
Transcription of broadcast news is an interesting and challenging application for large-vocabulary continuous speech recognition (LVCSR). We present in detail the structure of a manually segmented and annotated corpus including over 160 hours of German broadcast news, and propose it as an evaluation framework of LVCSR systems. We show our own experimental results on the corpus, achieved with a state-of-the-art LVCSR decoder, measuring the effect of different feature sets and decoding parameters, and thereby demonstrate that real-time decoding of our test set is feasible on a desktop PC at 9.2\,\% word error rate.
\end{abstract}
\noindent{\bf Index Terms}: LVCSR, corpus, German, broadcast news, real-time

\section{Introduction}

Transcription of broadcast news has been an active topic of research in the last decade. Various languages apart from English \cite{Beyerlein2002lvc,Gales2006pit} have been covered, including Arabic \cite{Lamel2007iam}, French \cite{Gauvain2005waw}, German  \cite{Rigoll2001tas,McTait2003t3l}, and Mandarin \cite{Yu2004tir,Sinha2006tcb}. The process of full transcription is usually divided into a segmentation and speech recognition step: first, the audio signal is segmented into speech and non-speech parts (e.\,g.\ music), and then the speech parts are decoded by a speech recogniser \cite{Gauvain2002tlb}.

In this paper, we do not aim at a full-featured transcription system, but rather focus on the challenge of speaker-independent German large-vocabulary continuous speech recognition (LVCSR), using a broadcast news speech corpus for evaluation.
This corpus was originally created at Duisburg University in the context of the ALERT project, and used as a whole for training a broadcast news transcription system \cite{Rigoll2001tas}. In that study the test set was rather small and not clearly specified. In contrast, this paper proposes a suitable corpus structure for systematic tuning and evaluation of a German LVCSR system, including subdivision into training, development, and test sets. We believe that provision of the corpus will be a valuable contribution for the engineering of speech recognition tools for diverse applications in the future, since recognition of German broadcast news is a considerably challenging application for LVCSR: on the one hand, recognition of broadcast news has to utilise large dictionaries to avoid frequent out-of-vocabulary events; on the other hand, it has to deal with the pecularities of the German language, such as frequent inflection (e.\,g.\ German {\em einen} or {\em einem} = English {\em a(n)}) and complex compound words (e.\,g.\ German {\em Spracherkennungsproblem} = English {\em speech recognition problem}) \cite{McTait2003t3l}.

Our work focused on structuring and preparing the German speech data available from the ALERT project into a corpus suitable for evaluation of speaker-independent LVCSR systems (Section \ref{sec:BCNCorpus}), assessing the benefit of  different acoustic feature sets (Section \ref{sec:FEx}), and fine-tuning of the decoding parameters (Section \ref{sec:Decoding}) with respect to the trade-off between real-time factor (RTF) and word error rate (WER). We will present the results of our experiments in Section \ref{sec:Results}, and show that real-time processing on a state-of-the-art PC is feasible at a WER of less than 10\,\%, which is lower than the ones typically found in previous works on transcription of German broadcast news.

\section{The Broadcast News Corpus}
\label{sec:BCNCorpus}

\begin{figure}[ht]
\begin{center}
\includegraphics[width=0.6\columnwidth]{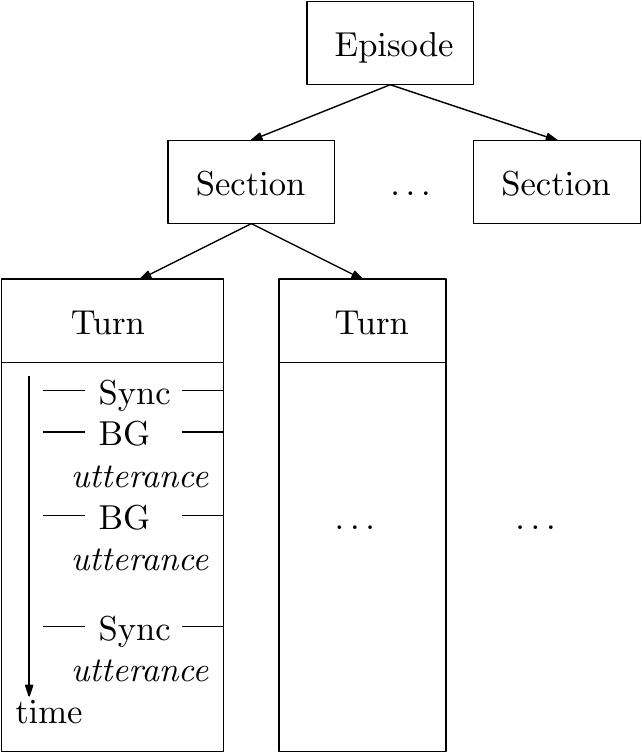}
\end{center}
\vspace{-0.3cm}
\caption{
\em Hierarchical structure of the XML format. It allows a flexible annotation and segment extraction from audio files ({\em Episode}) by providing {\em Sections} corresponding e.\,g.\ to speech or music parts, subdivided into {\em Turns} of a certain speaker, which consist of a sequence of utterances for which time alignments ({\em Sync}) are provided. Changes in background ({\em BG}), such as noise conditions are also annotated.
}
\vspace{0.3cm}
\label{fig:Trans}
\end{figure}

The Broadcast News (BCN) Corpus was recorded in 2000--01 at Duisburg University and consists of over 160 hours of German speech, including mainly radio broadcasts, but also news on television ({\em Tagesschau}). For the television parts, video is available as an additional modality, e.\,g.\ for audiovisual speech recognition, but was not considered in our experiments. Audio is sampled at 16\,kHz with 16-bit PCM encoding. The audio files each correspond to a broadcast with several minutes length. 
The segmentation and transcription is contained in a flexible XML file format that is illustrated in Figure \ref{fig:Trans}:
first, each of the audio files is coarsely divided into {\em sections}, corresponding e.\,g.\ to jingles at the beginning of the broadcast, or speech parts. Each speech part is segmented into {\em turns} of different speakers which are listed at the beginning of the file. Gender information could be used to train specific acoustic models, which we leave as a topic for further experiments. 
Each turn corresponds to one or several news messages read by one speaker and is transcribed on the word level, using reference texts from the web sites of the radio and TV stations whenever available. Manually created alignments ({\em sync times}) divide the turns into {\em utterances}, which usually correspond to complete news messages and hence contain a few sentences. The average utterance length in the corpus is 80 words. %
Alignments are also provided whenever the background condition (clean speech, music, background noise, or a combination of them) changes, which allows further partitioning of the utterances into segments with only a specified background condition. Finally, signal segments with low speech quality -- due to transmission or microphone problems -- are also identified.
We divided the utterances from the clean speech parts into a training, development, and test set, ignoring the low-quality segments (approx.\ 1.4\,\% of the total utterance length). %
We realised the constraint of speaker-independency by choosing these sets such that no radio or TV channel occurs in more than one set. The recording length and number of utterances for the three sets is shown in Table \ref{tab:Corpus}, while the number of out-of-vocabulary (OOV) entries is outlined in Table~\ref{tab:CorpusVocab} for a 148\,K language model vocabulary size.

\begin{table}[ht]
    \begin{center}
    \begin{tabular}{|l|l|l|r|r|}
    \hline
    \bf Channel & \bf Type & \bf Year(s) & \bf h & \bf \# Words \\
    \hline \hline
    \multicolumn{5}{|l|}{\bf Training set} \\
    \hline
    DLF & Radio &  2000--01 & 67.81 & 535\,K \\
    SWR2 & Radio & 2000 & 38.42 & 333\,K \\
    SWR2 Info & Radio & 2000 & 1.88 & 18\,K \\
    WDR2 & Radio & 2000 & 26.75 & 224\,K \\
    NDR4 & Radio & 2000 & 11.41 & 98\,K \\
    \hline
    {\bf total} &  &    & 146.28 & 1\,209\,K \\
    \hline
    \hline
    \multicolumn{5}{|l|}{\bf Development set} \\
    \hline
    ARD & TV & 2000--01 & 8.93 & 84\,K \\
    \hline
    \hline
    \multicolumn{5}{|l|}{\bf Test set} \\
    \hline
    DW & Radio & 2000--01 & 6.33 & 50\,K \\
    \hline
    \end{tabular}
    \end{center}
    \vspace{-0.3cm}
\caption{
\em Overview over the training, development, and test set of the BCN corpus. ``h'' indicates total utterance length in hours. Note that for our experiments we only used the audio channel of the TV data.
}
\label{tab:Corpus}
\end{table}

\begin{table}[ht]
\vspace{0.3cm}
\begin{center}
\begin{tabular}{|l|r|r|r|}
\hline
{\bf Set} & {\bf \# OOV} & {\bf \% OOV} \\
\hline
\hline
{\bf Training} &    3\,585 & 0.55 \\
\hline
{\bf Development} & 734 & 1.11 \\
{\bf Test} &        127 & 0.36 \\
\hline
\end{tabular}
\end{center}
\vspace{-0.3cm}
\caption{
\em Number of unique out-of-vocabulary entries {\em (\# OOV)}, and out-of-vocabulary rate {\em (\% OOV)}, related to the 148\,K language model vocabulary size, in the training, development, and test set of the BCN corpus. Note that we handle OOV words in the training set by mapping to a garbage model.
}
\label{tab:CorpusVocab}
\end{table}

\section{Feature Extraction}
\label{sec:FEx}

To be able to measure how recognition performance is affected by different acoustic features, we extracted three feature sets. A baseline set contained Mel Frequency Cepstral Coefficients (MFCC) 1--12 and signal log-energy, and their first ($\delta$) and second order regression coefficients ($\delta\delta$), which were extracted using the openEAR feature extractor \cite{Eyben2009oit}, and are identical to the features extracted by the HTK toolkit \cite{Young2006thb}.

Next, this feature set was augmented by 12 Line Spectral Pairs (LSP), along with their $\delta$ and $\delta\delta$ coefficients, computed with openEAR. To reduce the number of parameters that had to be estimated for the acoustic models, and to have feature sets of comparable dimension, we applied Principal Components Analysis (PCA) to the 75-dimensional feature vectors. For PCA, we computed the covariance matrix of the training set when testing with the development set, or the one of the union of training and development set when testing with the test set. The feature space was reduced by considering only the first $N$ eigenvalues (in descending magnitude) $\lambda_1, \dots, \lambda_N$ such that
\begin{equation}
\frac{\sum_{i=1}^N \lambda_i}{\sum_{i=1}^{75} \lambda_i} \geq 0.9999 ,
\end{equation}
and choosing the corresponding eigenvectors as basis. For both cases (training set or union of training and development set), the resulting dimensionality was $N=38$. The final features will be denoted by MFCC-LSP-PCA.

Finally, as a third feature set, we chose the Perceptual Linear Prediction (PLP) coefficients 1--12 along with log-energy and their $\delta$ and $\delta\delta$ coefficients, computed with HTK \cite{Young2006thb}.

\section{Acoustic and Language Models}
\label{sec:Decoding}

In contrast to previous work on German broadcast data \cite{Rigoll2001tas} using the ``Ducoder'' stack decoder \cite{Willett2000dtd}, we implemented a LVCSR system on top of HDecode \cite{Young2006thb}. Our acoustic models included tied-state Hidden Markov Models (HMMs) with three emitting states, corresponding to decision-tree clustered cross-word triphones built from 54 monophone models. Apart from 51 models for the common German phoneme alphabet and silence, the monophones also included a model for the nasal {\em en} vowel that quite frequently occurs in German due to a considerable number of French foreign words (e.\,g.\ {\em Arrangement}). Two additional models were needed for technical reasons: a `garbage' phoneme model was trained to model out-of-vocabulary (OOV) words in the training set, in order to be able to use all available utterances from the training set. Also, due to utterances containing multiple sentences in the corpus, it was necessary to introduce an additional `within-utterance' silence model (`full stop'), because the HDecode decoder assumes an end of turn once `standard' silence is detected.
After building the tied-state triphones, Gaussian mixtures (GM) were split until each state had 8 GMs, with an additional split for silence (16 GMs).

As has been shown in \cite{Rigoll2001tas}, using a domain-specific  language model (LM) can vastly increase recognition rate. For our experiments, we built a trigram LM from the archive of the German newspaper {\em taz (die tageszeitung)}, consisting of 633\,611 articles from the years 1986-2000 with 185.9 million words in total. For the 148.5\,K LM vocabulary, pronunciations were taken from a semi-automatically generated dictionary. Notably, for lower vocabulary sizes the OOV rate increased significantly (see Table \ref{tab:OOVVocab}).
To match the structure of the utterance annotation appropriately with the LM, we replaced the full stops in the articles by a word corresponding to the aforementioned `within-utterance' silence model. We ended up with an LM size of approx.\ 400 megabytes in ARPA format, hence the LM fits well into the memory of a state-of-the-art PC, and for our purpose no special effort to cope with memory requirements was needed, unlike in \cite{Willett2000dtd}.
Table \ref{tab:Perplexity} shows the perplexity -- measured by the method described in \cite{Young2006thb} -- of the transcription of the training, development, and test sets, and the BCN corpus as a whole, using the unigram, bigram, and trigram LM.

\begin{table}
\begin{center}
\begin{tabular}{|l|c|c|c|c|c|}
\hline
{\bf $n$ words} & 10\,K & 20\,K & 50\,K & 100\,K & 148\,K \\
\hline
{\bf \% OOV} & 14.1 & 9.3 & 4.7 & 2.1 & 0.36 \\
\hline
\end{tabular}
\end{center}
\vspace{-0.3cm}
\caption{\em OOV rates over the BCN test set when using only the $n$ most frequent words in the texts for building the language model.}
\label{tab:OOVVocab}
\end{table}

\begin{table}
\begin{center}
\begin{tabular}{|l|c|c|c|}
\hline
{\bf Set} & {\bf Unigram} & {\bf Bigram} & {\bf Trigram} \\
\hline
\hline
{\bf Training}    & 2\,487 & 462 & 299 \\
{\bf Development} & 2\,150 & 445 & 291 \\
{\bf Test}        & 2\,751 & 473 & 302 \\
\hline
{\bf Total}       & 2\,474 & 462 & 299 \\
\hline
\end{tabular}
\end{center}
\vspace{-0.3cm}
\caption{\em Perplexity of the training, development, and test set transcriptions, measured using a language model built from 633\,611 articles (185.9\,M words) from the German newspaper {\em taz}.}
\label{tab:Perplexity}
\end{table}

\section{Experimental Results}
\label{sec:Results}

\begin{figure}[t]
\includegraphics[width=\columnwidth]{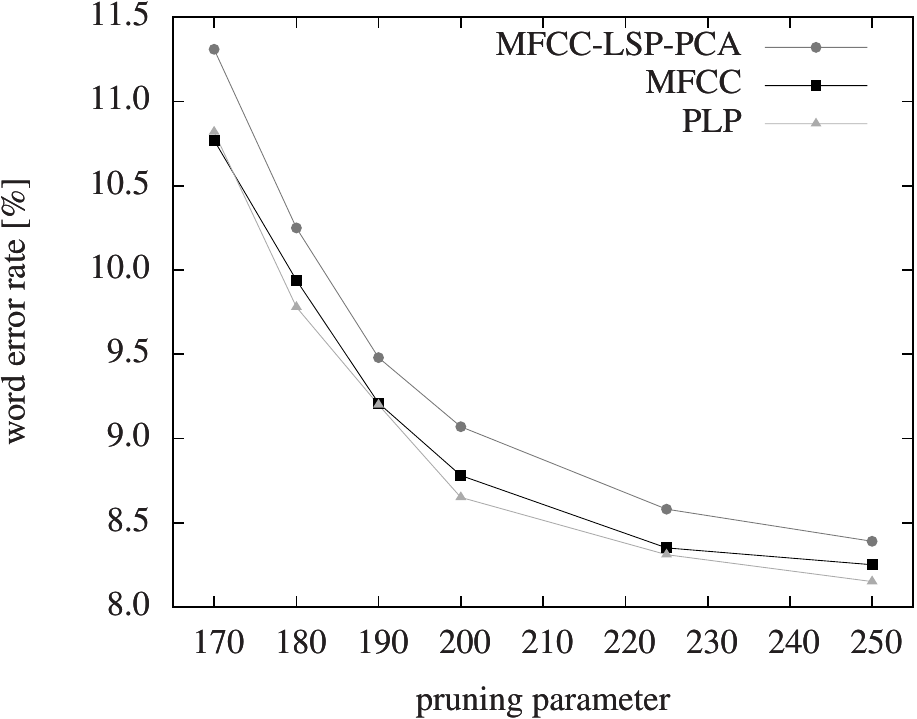}
\caption{\em Effect of beam width on word error rate for the test set of the BCN corpus and three different feature sets.}
\label{fig:PruningWER}
\end{figure}

We defined suitable ranges for the decoding parameters on the development set: the word insertion penalty used by HDecode was set to a log probability of -4.0, the HDecode beam width was chosen between 170.0 and 250.0, and values of the LM scale factor between 12.0 and 20.0. Then, we joined the training and development sets and evaluated different combinations of these parameters on the test set.
First, Figure \ref{fig:PruningWER} shows the WER for the three different feature sets (Section \ref{sec:FEx}) at various beam widths of the decoder (170.0, 180.0, 190.0, 200.0, 225.0, and 250.0). With the minimal pruning (beam width 250.0) and PLP features, a WER of 8.1\,\% can be achieved. Comparing the results with different feature sets, it can be seen that while a WER of below 10\,\% is feasible for all three feature sets, the overall picture clearly shows that PLP outperform MFCC and the combination of MFCC and LSP reduced by PCA. The latter exhibit the lowest performance in the test scenario.

\begin{figure}[t]
\includegraphics[width=\columnwidth]{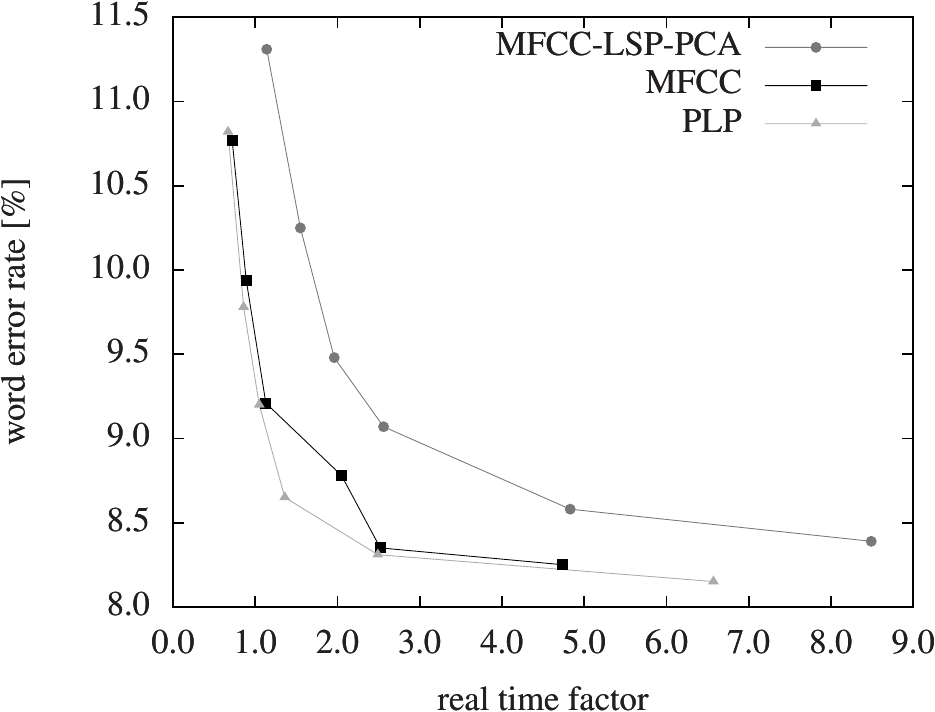}
\caption{\em Trade-off between real-time factor and word error rate for the test set of the BCN corpus when varying the beam width and feature set. The language model scale factor was 15.0. Measurements were carried out on a 3.0\,GHz desktop PC.}
\label{fig:RTFWER}
\end{figure}

Next, Figure \ref{fig:RTFWER} shows the trade-off between WER and RTF, again for the three different feature sets. It is remarkable that the 39 MFCC and PLP features are approximately on par concerning RTF, while the 38 PCA features from MFCC and LSP significantly increase the RTF needed for a given WER. Note that in this figure, different RTFs resulted from using the various aforementioned pruning settings, and were computed as the quotient of the total CPU time used by HDecode on the test set and the total utterance length of the test set. The measurement of CPU time was carried out with HTK 3.4 on a Linux PC with a 3.0\,GHz AMD Phenom II X4 processor and 8\,GB of RAM. Only one of the four cores was used for each decoding process. For the results in Figure \ref{fig:PruningWER} and \ref{fig:RTFWER}, the LM scale factor was 15.0.

\begin{figure}[ht]
\includegraphics[width=\columnwidth]{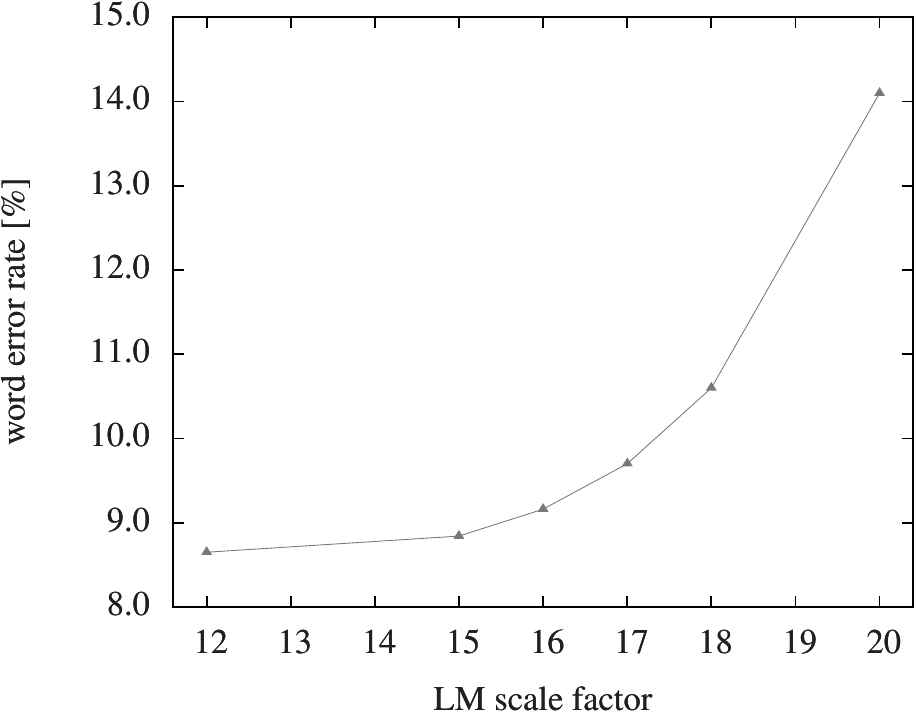}
\caption{\em Effect of language model scale factor on word error rate for the test set of the BCN corpus. A beam width of 200.0 and PLP features were used.}
\label{fig:LMWER}
\end{figure}

\begin{figure}[h!t]
\includegraphics[width=\columnwidth]{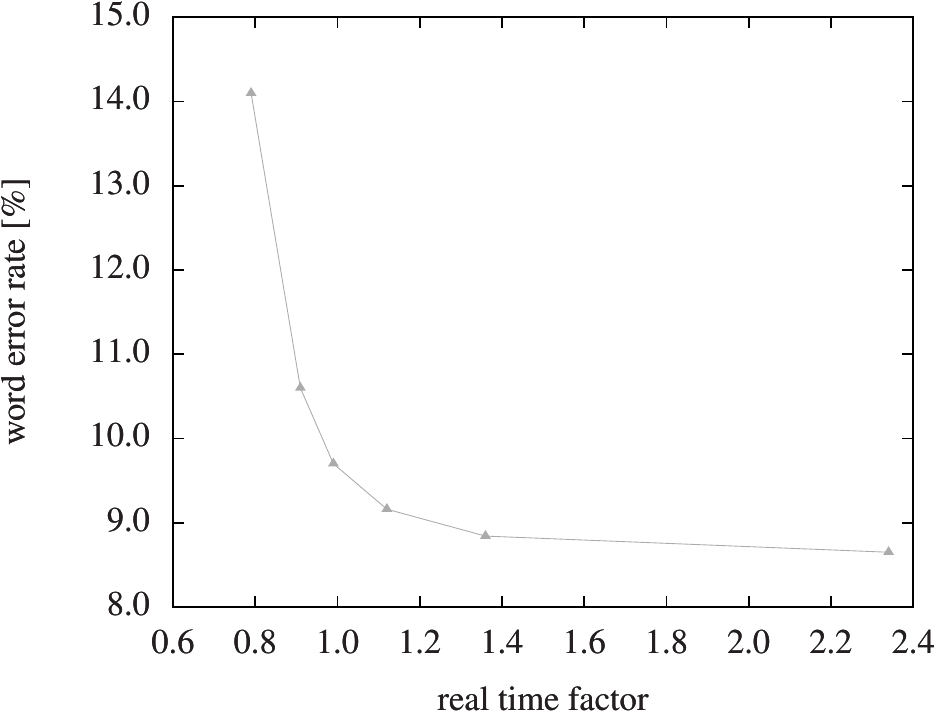}
\caption{\em Trade-off between real-time factor (on a 3.0\,GHz PC) and word error rate for the test set of the BCN corpus when varying the language model scale factor at a beam width of 200.0. PLP features were used. Comparing to Figure \ref{fig:RTFWER}, it can be seen that variation of the beam width generally delivers a lower word error rate for a given real-time factor.}
\label{fig:RTFWER-LM}
\end{figure}

Variation of the LM scale factor is another method to fine-tune the trade-off between WER and RTF: Figure \ref{fig:LMWER} depicts the WER induced by different scale factors, while Figure \ref{fig:RTFWER-LM} shows the trade-off between WER and RTF in analogy to Figure \ref{fig:RTFWER}, but varying the LM scale factor instead of the beam width. Here, PLP features were used. Opposing Figure \ref{fig:RTFWER-LM} and Figure \ref{fig:RTFWER}, one can see that variation of the beam width at a constant LM scale factor of 15.0 is more beneficial for improving the trade-off between WER and RTF than keeping the beam width constant (200.0) and varying the LM scale factor. In the latter case, a higher WER of 9.7\,\% has to be accepted for real-time decoding (with an LM scale factor of 17.0), compared to a WER of about 9.2\,\% achieved by a beam width of 190.0. Furthermore, a RTF of 0.67 at a WER of 10.82\,\% can be achieved by a beam width of 170.0, while accomplishing a RTF of 0.79 by a LM scale factor of 20.0 induces a WER of 14.10\,\%.

As a final note concerning WER, it is noteworthy that sometimes calculation of WER for German speech takes into account equivalence classes for compound words \cite{McTait2003t3l}. That is, if the reference transcription contains the word {\em Sportveranstaltung} ({\em sports event}) and the decoder splits it up into the sequence {\em Sport Veranstaltung}, this is {\em not} considered erroneous (more precisely, as one insertion and one substitution). However, we calculate WER without using such equivalence classes, for two reasons: first, we believe that the decoder should produce an orthographically correct transcription, which {\em Sport Veranstaltung} is not; second, it is expected that compound words are pronounced without pauses, which the system should be able to detect. The `full stop' word (Section \ref{sec:Decoding}) was however ignored in computation of accuracy.

\section{Conclusions}

We presented a corpus of German broadcast news for fine-tuning and evaluation of German LVCSR systems, focusing on the decoding of clean speech. Using a decoding system based on state-of-the-art LVCSR research and a LM vocabulary size of 148\,K, we achieved a word error rate of about 9\,\% in real-time on a desktop PC. Different feature sets were evaluated, whereby PLP coefficients showed the best performance. Furthermore, we extensively fine-tuned the decoding parameters, and identified the beam widths and LM scale factors needed to achieve various real-time factors.
Comparing our results to previous work on broadcast news transcription, we conclude that our results are competitive, especially given that German is a considerably challenging language for LVCSR applications. Naturally, comparisons with previous studies must be done very carefully, as our results were achieved on manually segmented data, and capabilities of PC hardware have developed at a fast pace in the last decade.

In the future, we will use the BCN corpus to evaluate novel hybrid decoders and approaches for feature extraction, e.\,g.\ decoders using bidirectional long short-term memory recurrent neural networks or features generated by non-negative matrix factorisation. While both of these have been successfully used in whole-word recognition tasks \cite{Schuller2010nmf}, it will be an interesting research topic to assess their usefulness for German LVCSR.

\section{Acknowledgement}
The research leading to these results has received funding from the European Community's Seventh Framework Programme (FP7/2007-2013) under grant agreement No.\ 211486 (SEMAINE).
\\
The authors would like to thank Uri Iurgel, Andreas
Kosmala, Steffen Werner, and Daniel Willett for their
contributions within the ALERT project during their period at the Department of Computer Science at Gerhard-Mercator-University Duisburg.

\eightpt
\bibliographystyle{IEEEtran}
\bibliography{is2010}

\end{document}